\pdfoutput=1

\documentclass[11pt]{article}

\usepackage[final]{acl}

\usepackage{times}
\usepackage{latexsym}

\usepackage[T1]{fontenc}

\usepackage[utf8]{inputenc}

\usepackage{microtype}

\usepackage{inconsolata}

\usepackage{graphicx}

\usepackage{enumitem}

\usepackage{multirow}

\usepackage{tabularx}

\usepackage{makecell}

\usepackage{colortbl}

\usepackage{graphicx}

\usepackage{booktabs}

\usepackage{amsmath}

\usepackage{algorithm}

\usepackage{algorithmic}

\usepackage{float}

\usepackage{tcolorbox}

\usepackage{ragged2e}

\usepackage{changepage}

\usepackage{fontawesome}

\title{Graph Counselor: Adaptive Graph Exploration via Multi-Agent Synergy to Enhance LLM Reasoning}

\author{
  Junqi Gao\textsuperscript{1,2,}\thanks{This work was done during his internship at Shanghai Artificial Intelligence Laboratory.},
  Xiang Zou\textsuperscript{2},
  Ying Ai\textsuperscript{3},
  Dong Li\textsuperscript{1,2,}\thanks{Corresponding authors: Dong Li and Biqing Qi.},
  Yichen Niu\textsuperscript{3},
  Biqing Qi\textsuperscript{1,}$^\dag$, 
  Jianxing Liu\textsuperscript{3} \\
  $^1$ Shanghai Artificial Intelligence Laboratory \\
  $^2$ School of Mathematics, Harbin Institute of Technology \\
  $^3$ Department of Control Science and Engineering, Harbin Institute of Technology \\
  {\tt\small \{gjunqi97,arvinlee826,qibiqing7\}@gmail.com,} \\ 
  {\tt\small \{24s004010,24s112113,niu62\}@stu.hit.edu.cn, \{jx.liu\}@hit.edu.cn}
  }

\begin{document}
\maketitle
\begin{abstract}

Graph Retrieval Augmented Generation (GraphRAG) effectively enhances external knowledge integration capabilities by explicitly modeling knowledge relationships, thereby improving the factual accuracy and generation quality of Large Language Models (LLMs) in specialized domains. However, existing methods suffer from two inherent limitations: 1) \textbf{Inefficient Information Aggregation}: They rely on a single agent and fixed iterative patterns, making it difficult to adaptively capture multi-level textual, structural, and degree information within graph data. 2) \textbf{Rigid Reasoning Mechanism}: They employ preset reasoning schemes, which cannot dynamically adjust reasoning depth nor achieve precise semantic correction. To overcome these limitations, we propose Graph Counselor, an GraphRAG method based on multi-agent collaboration. This method uses the Adaptive Graph Information Extraction Module (AGIEM), where Planning, Thought, and Execution Agents work together to precisely model complex graph structures and dynamically adjust information extraction strategies, addressing the challenges of multi-level dependency modeling and adaptive reasoning depth. Additionally, the Self-Reflection with Multiple Perspectives (SR) module improves the accuracy and semantic consistency of reasoning results through self-reflection and backward reasoning mechanisms. Experiments demonstrate that Graph Counselor outperforms existing methods in multiple graph reasoning tasks, exhibiting higher reasoning accuracy and generalization ability.
Our code is available at \faGithub~\href{https://github.com/gjq100/Graph-Counselor.git}{\color{blue!50}{Graph-Counselor}}.
\end{abstract}

\section{Introduction}
Large Language Models (LLMs) are revolutionizing Natural Language Processing (NLP), demonstrating remarkable capabilities in tasks, such as text comprehension, explanation, and generation (\citealp{schaeffer2023are}; \citealp{qi2023large}). However, the issue of "hallucination", where LLMs generate factually inaccurate content, remains a critical challenge, especially in specialized domains \citep{rawte2023survey}. To mitigate this issue, incorporating external knowledge has emerged as a key strategy for enhancing LLM reliability (\citealp{gao2023retrieval}; \citealp{qi2024interactive}).

Retrieval-Augmented Generation (RAG) improves factual consistency by retrieving relevant information from external textual corpora (\citealp{asai2024selfrag}; \citealp{jeong2024adaptive}). However, conventional RAG primarily retrieves independent text units, which limits its ability to reason across multiple segments and integrate the structured knowledge embedded within them. In contrast, graph structures (e.g., knowledge graphs, KGs) offer a more systematic organization of knowledge, facilitating the construction of coherent knowledge chains to support deep reasoning \citep{liu2024lost}. Motivated by this, Graph Retrieval-Augmented Generation (GraphRAG) has been proposed to explicitly model knowledge relationships during retrieval, enabling LLMs to more accurately access and leverage structured knowledge (\citealp{edge2024local}; \citealp{wu2024medical}).

Current GraphRAG methods mainly follow two strategies. The first category is \textbf{retrieval-driven GraphRAG methods}, which first retrieve relevant information from KGs and then feed it to LLMs to generate answers. These methods rely on efficient retrievers, such as graph encoders (\citealp{mavromatis2024gnn}; \citealp{liu-etal-2024-knowledge-graph}) that explicitly model graph topology and node relationships or LLM fine-tuning (\citealp{chai2023graphllm}; \citealp{tang2024graphgpt}) that adapts language models to better interpret and query graph-organized knowledge. However, high computational costs and poor generalization have become bottlenecks. Moreover, the graph information retrieved by these methods are typically passed to LLMs in the form of linearized text (\citealp{mavromatis2024gnn}; \citealp{liu-etal-2024-knowledge-graph}) or code (\citealp{DBLP:journals/corr/abs-2408-13863}; \citealp{skianis2024graph}), which often results in the loss of critical structural information and weakens reasoning performance.

\begin{figure}[t]
  \includegraphics[width=\columnwidth]{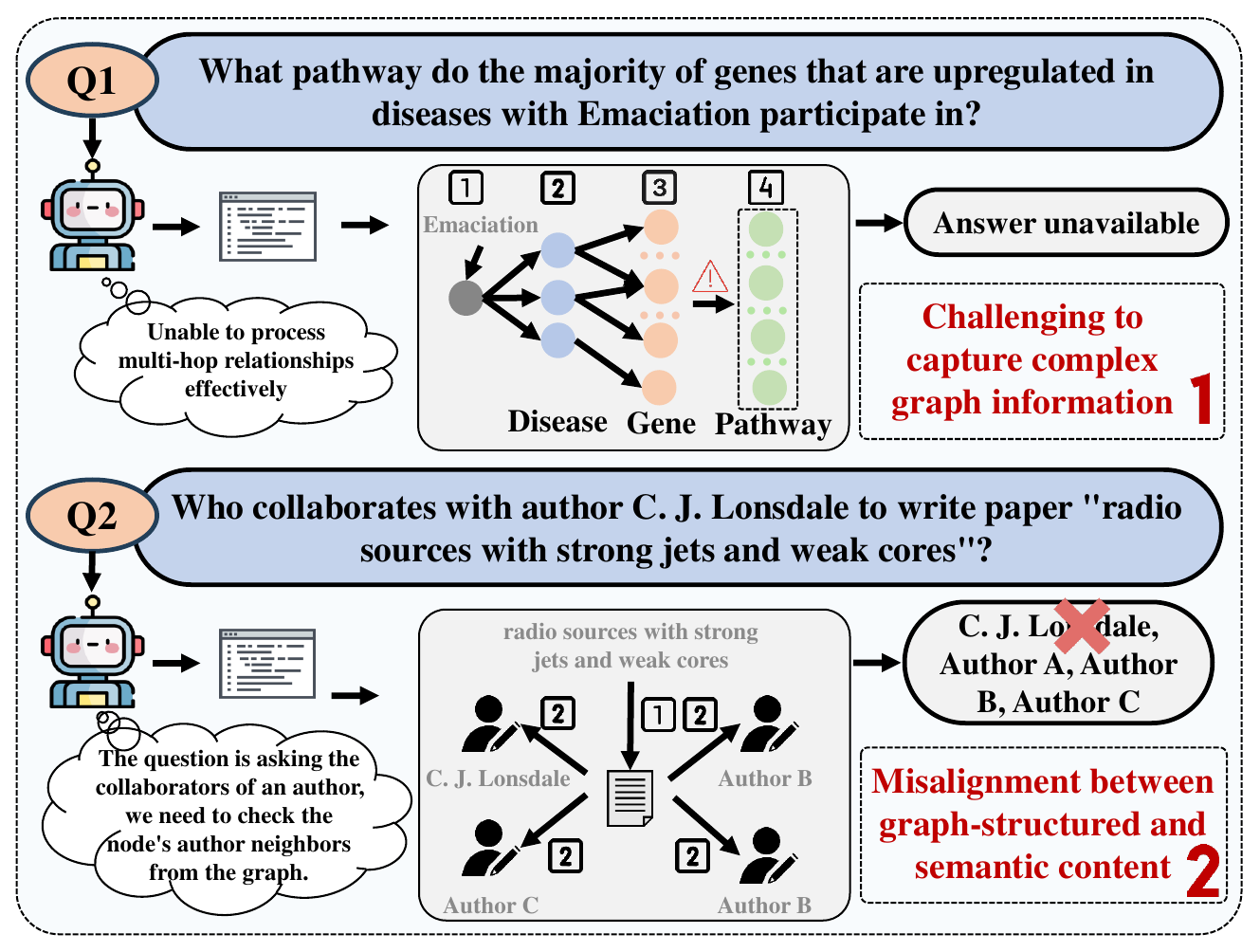}
  \caption{Two examples of LLM reasoning that highlight the two challenges.}
  \label{fig:reasoning_example}\vspace{-15pt}
\end{figure}

To avoid the aforementioned issues, \textbf{adaptive reasoning-based GraphRAG methods} have been proposed. These methods allow LLMs to interact with KGs in multiple rounds to dynamically adjust the information acquisition process, reducing the dependence on complex retrievers and additional training. Although this approach alleviates the problem of information loss to some extent, it still faces two major challenges: 
1) \textbf{Inefficient Information Aggregation}: Current methods (e.g., \citealp{DBLP:conf/acl/JinXZRZL0TWM024}; \citealp{markowitz-etal-2024-tree}; \citealp{luo2024graph}; \citealp{NEURIPS2024_4254e856}; \citealp{sunthink}; \citealp{jiang2023structgpt}) typically rely on a single agent to extract information in a fixed pattern, lacking the ability to adaptively model multi-level information in graph data. Specifically, these methods often employ a uniform information granularity when dealing with graph structures, only capable of capturing local text or simple topological relationships, and failing to effectively integrate multi-dimensional features such as node attributes, edge structures, and global degree information. This static aggregation mechanism not only limits the model's expressive power for complex graph structures but also leads to the neglect of key semantic information during transmission, reducing the overall efficiency and accuracy of the reasoning process. 2) \textbf{Rigid Reasoning Mechanism}: Most existing methods adopt preset reasoning paths and fixed reasoning iterations, unable to dynamically adjust the reasoning depth according to task complexity, resulting in "under-reasoning" or "over-reasoning" when facing problems of different difficulties. Meanwhile, due to the lack of an effective semantic alignment mechanism, the model is prone to deviating from the original query intention during reasoning, causing retrieval path bias and the introduction of erroneous information. Moreover, the inherent non-linear nature of graph structures and the linear understanding of text by language models create a natural gap (\citealp{choudhary2023complex}; \citealp{wang2024reasoning}), further exacerbating semantic understanding deviations and making it difficult to ensure the accuracy and consistency of reasoning results (Examples shown in Figure~\ref{fig:reasoning_example}).

To address these challenges, we propose Graph Counselor, a novel multi-agent collaborative reasoning framework that optimizes information extraction and improves reasoning accuracy through self-reflection. This framework introduces the Adaptive Graph Information Extraction Module (AGIEM), where three specialized agents collaborate in a hierarchical manner to extract and process graph information. The Planning Agent establishes a structured reasoning pathway by incorporating both query and contextual information, ensuring that the reasoning process follows a coherent logical sequence rather than haphazard retrieval. The Thought Agent refines the scope of information extraction by identifying the specific graph-related knowledge needed at current reasoning step, avoiding unnecessary retrieval from the entire graph. The Execution Agent dynamically adjusts retrieval strategies based on prior reasoning steps, ensuring that the extracted knowledge maintains its structural integrity and interdependencies. By iteratively operating within this framework, LLMs can more effectively leverage complex graph structures and improve reasoning precision.

In addition to optimizing information extraction, we introduce the Self-Reflection with Multiple Perspectives (SR) module to mitigate the misalignment between graph structures and semantic content. After AGIEM generates an answer, SR evaluates the reasoning path and final response for logical consistency, identifying potential errors or biases. If discrepancies are detected, SR summarizes key reasoning points, records error patterns, and adjusts input context accordingly. Furthermore, SR incorporates reverse reasoning and multi-perspective analysis to refine AGIEM’s understanding of queries and contextual information, leading to more reliable and semantically coherent reasoning outcomes.

In summary, our contributions are as follows:

$\bullet$ \textbf{Graph Counselor for enhanced GraphRAG reasoning}: We introduce a multi-agent framework that improves graph-based information retrieval and reasoning accuracy.

$\bullet$ \textbf{Adaptive Graph Information Extraction Module (AGIEM) for structured reasoning}: AGIEM uses a three-agent strategy (Planning, Thought, and Execution) to dynamically adapt, and effectively model complex graph structures and multi-level dependencies.

$\bullet$ \textbf{Self-Reflection with Multiple Perspectives (SR) for improved reasoning reliability}: SR corrects reasoning errors by incorporating self-reflection, reverse reasoning, and multi-perspective evaluation, enhancing reasoning reliability.

$\bullet$ \textbf{Multi-Dataset Empirical Validation}: Experiments on various benchmarks show that Graph Counselor outperforms existing methods in reasoning accuracy and generalization.

\section{Graph Counselor}

\subsection{Overview}

\begin{figure*}[t]
  \includegraphics[width=1\linewidth]{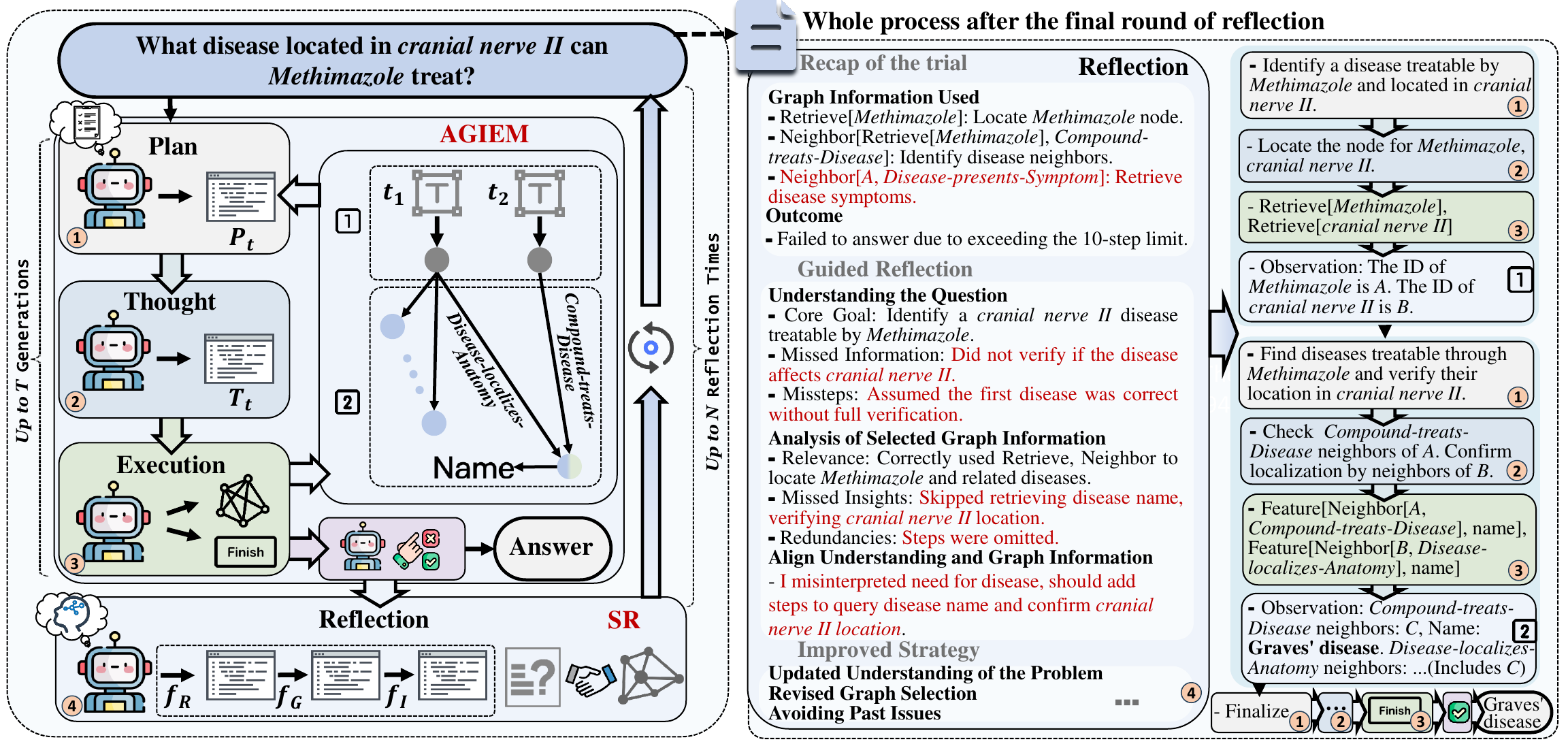}
  \caption{The workflow of Graph Counselor (left), with an example of reasoning process after reflection (right), where red highlights indicate errors and key reflections. The numbers in the boxes (left) correspond to the numbered reasoning steps shown in the process (right).}
  \label{fig:Graph Counselor}
  \vspace{-10pt}
\end{figure*}

Graph Counselor leverages multi-agent collaboration to flexibly extract graph structure information and optimize the inference mechanism, thereby improving the performance of LLMs on Knowledge Graph Question Answering (KGQA) tasks. Its workflow is illustrated in Figure~\ref{fig:Graph Counselor}. The system consists of two key modules: \textbf{1) Adaptive Graph Information Extraction Module (AGIEM)}, which utilizes a tri-agent collaboration strategy involving planning, reasoning, and execution to hierarchically parse and extract graph information, providing crucial support for the complex graph-related information required during the inference process. \textbf{2) Self-Reflection with Multiple Perspectives (SR)}, which defines a self-reflection mechanism for LLMs based on memory information. Through multi-perspective guidance, SR enhances the model's comprehension abilities, corrects biases in AGIEM’s query and context understanding, and offers improvement suggestions for subsequent reasoning.

\subsection{Adaptive Graph Information Extraction Module} 
\label{subsec:AGIEM}

\textbf{Graph Definition.} \,Let $\mathcal{G}=(\mathcal{V},\mathcal{E})$ be a KG. $\mathcal{V}$ and $\mathcal{E}$ denote the sets of nodes and edges, respectively. Each node $v \in \mathcal{V}$ comprises a unique identifier $I_v$ and a set of features, with each feature corresponding to a specific feature type. (e.g., "1047566": {"features": {"title": "Hand in Glove", "description": "", "price": "", "category": "books"}}). Each edge $r \in \mathcal{E}$ is represented by a label (e.g., "also-bought-item"). In this work, AGIEM performs reasoning on KGs in different domains.

\noindent\textbf{Planning Agent.} \,Given a question or the context from previous reasoning, Planning Agent first analyzes its meaning, identifies information relevant to inferring the correct answer, and then formulates the subsequent reasoning paths or determines that the query can already be inferred from the acquired graph information. For example, in Figure~\ref{fig:Graph Counselor}, given the question “What disease located in cranial nerve II can Methimazole treat?”, Planning Agent is expected to reason that “We need to identify a disease that is treatable by Methimazole and located in cranial nerve II.”

\noindent\textbf{Thought Agent.} \,Based on the reasoning results from Planning Agent and the query target, Thought Agent determines what graph information each step of the reasoning path needs, or analyzes the existing information to infer the query answer. In the given example, Thought Agent is expected to reason that “We need to locate the nodes for Methimazole and cranial nerve II in the graph first.” The collaboration between the Planning Agent and the Thought Agent clarifies the specific requirements for graph structure information in the process of multi-step reasoning to answer the question.

\noindent\textbf{Execution Agent.} \,Based on the previous reasoning results, we enable LLMs to adaptively extract graph-structured information to meet the needs for complex graph structures. To achieve this, inspired by \citep{DBLP:conf/acl/JinXZRZL0TWM024}, we have defined a diverse set of graph feature extraction components.
\begin{itemize}
    \item $\mathrm{Retrieve}(t)\rightarrow I_v$: It takes a query text $t$ as input and performs similarity search to retrieve the identifier $I_v$ of the node ($v \in \mathcal{V}$) that is most semantically relevant within the graph. This allows the localization of relevant nodes based on query semantics. (e.g., $\mathrm{Retrieve}(\texttt{Hand in Glove}) \rightarrow 1047566$)
    \item $\mathrm{Feature}({I}_v,\mathcal{T}_v)\rightarrow f_{vt}$: It takes a specific node identifier $I_v$ and a feature type $\mathcal{T}_v$ as input, returning the corresponding feature value $f_{vt}$. This extracts semantic information based on specific feature attributes of the node. (e.g., $\mathrm{Feature}(1047566, \texttt{category}) \rightarrow \texttt{books}$)
    \item $\mathrm{Neighbor}(I_v,r)\rightarrow I_{v'}$: It takes a specific node identifier $I_v$ and an edge label $r$ as input, returning the identifiers of all neighbor nodes $v'$ connected to $v$ by relation $r$. This captures the local relational structure of graph, focusing on the relationships between nodes and the information of their neighbors. (e.g., $\mathrm{Neighbor}(203088, \texttt{also-bought-item}) \rightarrow 203010$)
    \item $\mathrm{Degree}(I_v,r)\rightarrow D_v$: It takes a specific node identifier $I_v$ and an edge label $r$ as input, returning the count of neighbors related to $v$ by $r$. This component captures an important graph property: degree, which is a key query target and reflects the significance of nodes within the graph. (e.g., $\mathrm{Degree}(203088, \texttt{also-bought-item}) \rightarrow 1$)
\end{itemize}

We enable the Execution Agent to self-organize these functional components, not only allowing parallel extraction of multiple graph-structured information, but also permitting the combination of the components in series to meet the needs for complex graph information extraction. This is represented as follows:
\begin{equation}
\mathcal{X} = \left\{ \mathcal{P}_j(G)  \bigg | 
\begin{aligned}
    &\mathcal{P}_j = \mathcal{P}_{j1} \circ \cdots \circ \mathcal{P}_{jk}, \\
    &\mathcal{P}_{ji} \in \mathcal{F}, \, k \geq 1
\end{aligned}
\right\} \label{GKE}
\end{equation}

where $\mathcal{F}$ represents the set of four components: $\{\mathrm{Retrieve}(t),\  \mathrm{Feature}({I}_v,\mathcal{T}_v), \ \mathrm{Neighbor}(I_v,r), \ \\\mathrm{Degree}(I_v,r)\}$. $\circ$ denotes composition. (e.g., $\mathrm{Re}\text{-}\\ \mathrm{trieve}(t) \circ \mathrm{Feature}({I}_v,\mathcal{T}_v) = \mathrm{Feature}(\mathrm{Retrieve}(\\t), \mathcal{T}_v)$.

The Planning Agent, Thought Agent, and Execution Agent are executed collaboratively in sequence during each round of reasoning until the reasoning is completed or the specified iteration limit T is reached. This flexible graph knowledge extraction enables LLMs to efficiently perform complex graph-structured reasoning.

\subsection{Self-Reflection with Multiple Perspectives}
\label{subsec:SR}
A comprehensive understanding of both semantic and graph structural information is essential for LLMs reasoning. We propose SR, a mechanism that enhances the model’s reasoning process through multi-perspective reflection. Unlike traditional self-reflection methods, SR avoids over-reliance on previous decisions or inferences by exploring alternative, potentially more effective strategies. Additionally, SR refines the reasoning process by analyzing discrepancies between the graph structure information extracted by AGIEM and the semantic content of the queries. This enhances LLMs’ semantic understanding and dynamically updates the graph knowledge extraction strategy, ensuring better alignment between graph structure and semantic information.

At the core of SR is a multi-perspective reflection process, structured into three interrelated stages:

(1) \textbf{Recap \& Understanding}: The model revisits the current iteration’s queries and graph knowledge extraction process, identifying key reasoning objectives while reflecting on potential misunderstandings from multiple perspectives.

(2) \textbf{Analysis \& Adjustment}: The model analyzes potential omissions, redundancies, or inconsistencies in the reasoning process, particularly focusing on misalignments between graph structure and semantic information. This includes identifying missing or extraneous graph relationships and resolving conflicts in the reasoning path through adaptive adjustments.

(3) \textbf{Refinement \& Update}: Based on the reflection insights, the model refines its reasoning strategy to enhance subsequent steps, ensuring that graph structure and semantic information remain well-aligned.

By incorporating divergent thinking across multiple perspectives, SR effectively detects and corrects reasoning biases at different levels, guiding LLMs along the correct reasoning path. This process is crucial for improving the model’s performance in graph-based reasoning tasks. Prompts can be found in Appendix ~\ref{sec:appendixD2}.

\subsection{LLM State Transition Mechanism and Workflow}
We propose a structured workflow for LLMs in graph reasoning, as shown on the left side of Figure ~\ref{fig:Graph Counselor}. The inner layer is the multi-round iterative reasoning framework of AGIEM, while the outer layer incorporates a reflective architecture combined with SE.

\noindent\textbf{Inner-layer reasoning.} \,Based on different contexts, LLMs can function as a Planning Agent, Thought Agent, or Execution Agent, each performing its corresponding task. By optimizing contextual reasoning within a single LLM, we can achieve the effect of multi-agent collaboration. Within AGIEM’s multi-round iterative framework, LLMs adaptively switch agent roles. Relevant prompts can be found in Appendix ~\ref{sec:appendixD1}.

\noindent\textbf{Outer-layer reflection.} \,We introduce a judgment module where a reflection model provides a correctness flag based on the query and reasoning process. Related prompts are detailed in Appendix ~\ref{sec:appendixD3}. After AGIEM completes its reasoning, if the flag is not set to True and the predefined maximum number of reflections has not been reached, SE is executed. The reflection results are then updated in the inner-layer reasoning context, and AGIEM is re-executed until the reasoning result is judged as True or the reflection limit is reached. This approach ensures that SE is applied when necessary, improving the efficiency of the method.

The state transitions of LLMs facilitate the systematic integration of AGIEM and SE in Graph Counselor, thereby supporting complex graph reasoning tasks. A more detailed flow of the method is presented in the pseudocode in Appendix ~\ref{sec:appendixA}.

\section{Experiments}

\renewcommand{\arraystretch}{1.4}
\begin{table*}[!h]
  \caption{\label{tab-main overall performance}
    Model Performance(\%) on GRBENCH comparing Base LLMs, Text RAG, GraphRAG(1-hop and 2-hop), Graph-CoT, and Graph Counselor. We evaluate their performance based on Rouge-L (RL) and QwenScore (QS).
  }
\belowrulesep=0pt
\aboverulesep=0pt
  \centering
  \fontsize{8}{11}\selectfont
  \begin{tabularx}{\linewidth}{>{\centering\arraybackslash}p{1cm}>
  {\centering\arraybackslash}p{3.3cm}>{\centering\arraybackslash\columncolor{gray!10}}X>{\centering\arraybackslash}X>{\centering\arraybackslash\columncolor{gray!10}}X>{\centering\arraybackslash}X>{\centering\arraybackslash\columncolor{gray!10}}X>{\centering\arraybackslash}X>{\centering\arraybackslash\columncolor{gray!10}}X>{\centering\arraybackslash}X>{\centering\arraybackslash\columncolor{gray!10}}X>{\centering\arraybackslash}X}
    \toprule
    \multirow{2}{*}{} & \multirow{2}{*}{ \textbf{Model}} & \multicolumn{2}{c}{\textbf{Academic}} & \multicolumn{2}{c}{\textbf{E-commerce}} & \multicolumn{2}{c}{\textbf{Literature}} & \multicolumn{2}{c}{\textbf{Healthcare}} & \multicolumn{2}{c}{\textbf{Legal}}\\
    & &\textbf{RL} & \textbf{QS} &\textbf{RL} & \textbf{QS}& \textbf{RL} & \textbf{QS} &\textbf{RL} &\textbf{QS} &\textbf{RL} &\textbf{QS} \\
    \midrule
           \multirow{3}{*}{\rotatebox{90}{\textbf{Base}}} 
           & gemma-2-9b-it  & 9.57 & 9.13 & 12.05 & 9.00 & 7.82 & 13.33 & 7.42 & 5.56 & 15.67 & 11.67 \\
           & Mistral-NeMo-Instruct-2407  & 6.34 & 4.97 & 3.37 & 3.50 & 6.33 & 9.17 & 6.14 & 5.59 & 11.56 & 8.89\\
           & Llama-3.1-70B-Instruct  & 12.79 & 10.82 & 11.93 & 5.50 & 2.04 & 7.41 & 9.69 & 7.78 & 16.93 & 16.11\\
    \midrule
           \multirow{3}{*}{\rotatebox{90}{\makecell{\textbf{Text}\\ \textbf{RAG}}}} 
           & gemma-2-9b-it  & 9.67 & 9.10 & 19.19 & 17.00 & 13.56 & 15.83 & 4.57 & 3.70 & 30.05 & 29.44\\
           & Mistral-NeMo-Instruct-2407  & 7.22 & 5.65 & 13.33 & 10.50 & 9.68 & 11.25 & 4.33 & 2.96 & 21.73 & 19.44\\
           & Llama-3.1-70B-Instruct & 14.50 & 14.00 & 20.44 & 16.00 & 14.14 & 19.17 & 7.74 & 7.41 & 28.85 & 28.89\\
    \midrule
           \multirow{3}{*}{\rotatebox{90}{\makecell{\textbf{GraphRAG} \\ \textbf{(1-hop)}}}}
           & gemma-2-9b-it & 30.70 & 29.03 &  27.10 & 23.00 & 21.00 & 20.42 & 17.48 & 12.59 & 26.66 & 27.22\\
           & Mistral-NeMo-Instruct-2407  & 20.08 & 18.24 & 15.62 & 12.50 & 15.20 & 17.50 & 11.32 & 10.37 & 26.80 & 25.00\\
           & Llama-3.1-70B-Instruct  & 32.96 & 34.94 & 29.98 & 25.00 & 24.47 & 29.17 & 21.19 & 15.56 & 41.33 & 37.22\\
    \midrule
           \multirow{3}{*}{\rotatebox{90}{\makecell{\textbf{GraphRAG} \\ \textbf{(2-hop)}}}}
           & gemma-2-9b-it & 31.36 & 27.61 & 21.77 & 19.00 & 21.34 & 21.25 & 2.82 & 2.22 & 32.08 & 31.67\\
           & Mistral-NeMo-Instruct-2407  & 14.40 & 12.35 & 15.49 & 12.50 & 14.41 & 17.08 & 4.24 & 3.70 & 23.50 & 22.22\\
           & Llama-3.1-70B-Instruct  & 33.09 & 33.88 & 26.36 & 20.00 & 23.46 & 29.17 & 11.46 & 8.15 & 42.52 & 38.89\\
    \midrule
           \multirow{3}{*}{\rotatebox{90}{\makecell{\textbf{Graph}\\ \textbf{CoT}}}} 
           & gemma-2-9b-it  & 41.51 & 41.73 & 37.10 & 38.50 & 41.25 & 44.58 & 29.50 & 27.78 & 28.12 & 32.46\\
           & Mistral-NeMo-Instruct-2407  & 32.26 & 34.18 & 30.33 & 39.00 & 24.67 & 33.33 & 27.26 & 28.15 & 29.09 & 35.56\\
           & Llama-3.1-70B-Instruct  & 47.64 & 52.94 & 31.21 & 34.50 & 42.06 & 42.08 & 43.70 & 39.63 & 41.60 & 42.22\\
    \midrule
           \multirow{3}{*}{\rotatebox{90}{\makecell{\textbf{Graph} \\ \textbf{Counselor}}}} 
           & gemma-2-9b-it  & 55.58 & 54.07 & 49.02 & 50.50 & 55.41 & 57.08 & 42.21 & 37.41 & 35.74 & 38.89\\
           & Mistral-NeMo-Instruct-2407  & 54.15 & 53.37 & 44.46 & 44.00 & 47.71 & 53.33 & 43.87 & 37.41 & 53.35 & 52.78\\
           & Llama-3.1-70B-Instruct  & \textbf{60.11} & \textbf{63.29} &  \textbf{48.33} & \textbf{49.00} &  \textbf{56.31} & \textbf{57.08} &  \textbf{48.90} & \textbf{40.74} &  \textbf{53.84} & \textbf{54.44}\\
    \midrule
           & \textbf{Average Improvement} & \textcolor{red}{16.14\text{↑}} & \textcolor{red}{13.96\text{↑}} & \textcolor{red}{14.39\text{↑}} & \textcolor{red}{10.50\text{↑}} & \textcolor{red}{17.15\text{↑}} & \textcolor{red}{15.83\text{↑}} & \textcolor{red}{11.51\text{↑}} & \textcolor{red}{6.67\text{↑}} & \textcolor{red}{14.71\text{↑}} & \textcolor{red}{11.96\text{↑}} \\
    \bottomrule   
  \end{tabularx}
\end{table*}

\subsection{Experimental Setup}
\textbf{Dataset.} \,In this study, we used the \textbf{GRBENCH} dataset \citep{DBLP:conf/acl/JinXZRZL0TWM024} to assess the ability of LLMs to interact with external knowledge graphs. GRBENCH consists of 10 real-world graphs across five domains (\textbf{Academic}, \textbf{E-commerce}, \textbf{Literature}, \textbf{Healthcare}, and \textbf{Legal}) with 1,740 questions. These questions are divided into three difficulty levels: simple (700 questions, single-hop reasoning), medium (910 questions, multi-hop reasoning), and hard (130 questions, requiring inductive reasoning). The questions are designed to be answerable based on the information within the graphs, aiming to simulate real-world application scenarios in specific domains.\\
By conducting systematic experiments on \textbf{GRBENCH}, we are able to comprehensively assess the performance of the proposed Graph Counselor and analyze its effectiveness under different demonstration settings, base LLMs, and question difficulty levels. \\
\textbf{Baselines}. \,We compare our proposed Graph Counselor with the following three RAG-based methods.
\begin{itemize}
\item \textbf{LLMs}: To assess whether LLMs can answer domain-specific questions solely based on their internal knowledge without external data, we adopt a standard prompting strategy. This involves providing concise instructions to allow LLMs to autonomously generate answers.
\item \textbf{Text RAG} \citep{gao2023retrieval}: This method treats external graphs as textual corpora and employs a retriever to extract relevant information. The retrieved text is used as contextual input to enhance LLM performance in question answering.
\item \textbf{GraphRAG}: As an extension of Text RAG, GraphRAG linearizes both retrieved text or nodes and their associated subgraphs into textual sequences for contextual augmentation. In our main experiments, GraphRAG \citep{ye-etal-2024-language} retrieves 1-hop and 2-hop subgraphs. Graph-CoT \citep{DBLP:conf/acl/JinXZRZL0TWM024}, the current state-of-the-art variant of GraphRAG, employs iterative reasoning to incrementally gather critical information from the graph, mitigating information loss caused by excessively long contexts.
\end{itemize}
For all methods, we ensure generalizability by evaluating across six LLM backbones: Mixtral-8x7B-Instruct-v0.1 , Mistral-NeMo-Instruct-2407 \citep{mixtral2024}, Qwen2.5-7B-Instruct, Qwen2.5-72B-Instruct \citep{yang2024qwen2}, Llama-3.1-70B-Instruct \citep{dubey2024llama}, and Gemma-2-9b-it \citep{team2024gemma}.\\
\textbf{Evaluation Metrics}. \,To comprehensively evaluate the performance of methods, we categorize the evaluation metrics into rule-based and LLM-based measurements. For rule-based metrics, we select the Rouge-L (RL) metric, which calculates the ratio of the longest common subsequence between system outputs and ground truth answers relative to the length of reference texts. For LLM-based metrics, we employ Qwen2.5-72B-Instruct and Llama3.1-70B-Instruct to assess the consistency between generated responses and reference answers. The proportion of questions judged as correct by each LLM is calculated as QwenScore (QS) and LlamaScore (LS) respectively.\\
\textbf{Parameter Configuration}. \,For the retrieval model, we utilize Mpnet-v2 \citep{10.5555/3495724.3497138} with FAISS \citep{johnson2019billion} for efficient indexing. In Graph Counselor, we set the temperature to 0.7 and top-p sampling to 0.9 to encourage diverse reasoning outputs.

\subsection{Overall Performance}
The main experimental results are shown in Table \ref{tab-main overall performance}. From the results, we observe that: 1) Graph Counselor demonstrates a significant advantage over other methods, achieving up to a 24.2\% improvement in the R-L metric compared to the state-of-the-art GraphRAG approach. Additional results for other LLM backbones and detailed information on the LS metric can be found in Appendix \ref{sec:appendixB}; 2) The performance of GraphRAG when retrieving 2-hop subgraphs is not always superior to retrieving 1-hop subgraphs. This could be because 2-hop subgraphs contain more nodes and edges, which, while potentially providing richer semantic information, may also introduce a large amount of irrelevant or even distracting information, thereby affecting retrieval quality. Therefore, flexibly selecting whether to leverage graph structure information based on the task requirements can enhance the adaptability and performance of GraphRAG. This further supports the rationality and effectiveness of the Graph Counselor design; 3) Overall, GraphRAG outperforms TextRAG, however, its advantage is less pronounced on the Legal dataset. This may be due to the richer contextual information in the queries of the Legal dataset, allowing TextRAG to directly retrieve relevant text chunks based on key phrases, thereby reducing its reliance on reasoning over graph structures.

\subsection{Additional Comparative Experiment}
To further validate the generality of Graph Counselor, we conducted comparative experiments on an additional KGQA dataset, WebQSP \cite{yih2016value}. Here, we used RL, QS, and LS as metrics. As shown in Table \ref{tab:webqsp}, the experimental results indicate that Graph Counselor still maintained significantly higher optimal performance on WebQSP. Notably, when using Mistral-NeMo-2407 as base model, Graph Counselor achieved an $12.50\%$ improvement in QS, an $11.00\%$ improvement in LS, and an $8.99\%$ improvement in RL compared to the second-best method, Graph-CoT. This further confirms the broad effectiveness of Graph Counselor.

\begin{table}[h]
\centering
\renewcommand{\arraystretch}{1.0}
\caption{Performance Comparison of Different Methods on WebQSP Dataset.}
\label{tab:webqsp}
\scriptsize
\begin{tabular}{lllcc}
\toprule
\multirow{2}{*}{Model} & \multirow{2}{*}{Method} & \multicolumn{3}{c}{Metrics (\%)} \\
\cmidrule(lr){3-5}
 &  & QS & LS & RL \\
\midrule
\multirow{5}{*}{gemma-2-9b-it} 
 & Base & 49.00 & 56.00 & 31.76 \\
 & Text RAG & 49.50 & 56.00 & 32.08 \\
 & GraphRAG (1-hop) & 51.50 & 56.50 & 34.95 \\
 & Graph-CoT & 52.50 & 56.50 & 37.01 \\
 & Graph Counselor & \textbf{59.00} & \textbf{60.00} & \textbf{42.81} \\
\addlinespace
\midrule
\multirow{5}{*}{Mistral-NeMo-2407} 
 & Base & 47.00 & 51.00 & 30.17 \\
 & Text RAG & 47.00 & 52.00 & 30.49 \\
 & GraphRAG (1-hop) & 48.00 & 53.00 & 34.19 \\
 & Graph-CoT & 48.00 & 54.00 & 36.12 \\
 & Graph Counselor & \textbf{60.50} & \textbf{65.00} & \textbf{45.11} \\
\bottomrule
\end{tabular}
\vspace{-10pt}
\end{table}

\begin{figure*}[!h]
\includegraphics[width=\textwidth]{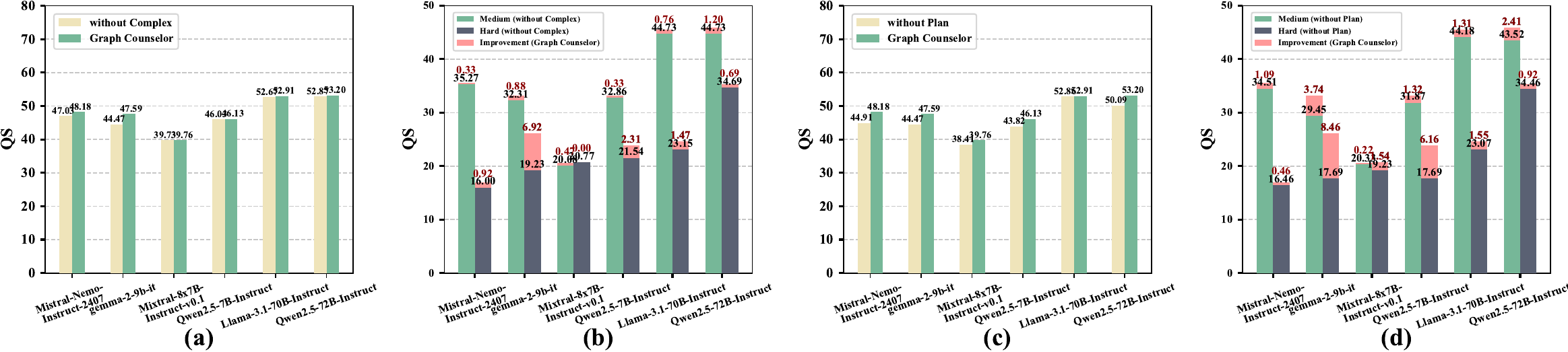}
  \caption{(a): Results comparison between without Complex and Graph Counselor on GRBENCH; (b): Results comparison between without Complex and Graph Counselor on GRBENCH with different levels; (c): Results comparison between without Plan and Graph Counselor on GRBENCH; (d): Results comparison between without Plan and Graph Counselor on GRBENCH with different levels.}
  \label{fig: without plan or complex}
  \vspace{-10pt}
\end{figure*}

\subsection{Ablation Study}
\subsubsection{How important is Plan and Complex Graph Information?}
To evaluate the impact of the Planning Agent and Execution Agent in the Graph Counselor framework, we ensure the reliability of our experimental results by keeping all other experimental settings—including datasets, evaluation metrics, and LLM backbones—identical to those in the main experiment. Based on the comparative analysis of the results shown in Figure~\ref{fig: without plan or complex}, we derive the following key findings: 1) The role of the Planning Agent in guiding inference paths: Removing the Planning Agent leads to an average decrease in accuracy of up to 6.1\% on medium- and high-difficulty questions, which is shown in (c) and (d). This result validates the effectiveness of the module in improving model performance on challenging problems through a dual mechanism of task decomposition and inference path planning. A possible reason for this improvement is that the Planning Agent decomposes complex questions into an ordered sequence of subtasks and generates structured inference pathways, which sequentially guide the system through knowledge retrieval, logical reasoning, and conclusion synthesis, thereby significantly enhancing inference capabilities; 2) The positive impact of the Execution Agent on complex reasoning: When the Execution Agent is limited to using a single component at a time, the accuracy on medium- and high-difficulty questions drops by up to 3.6\%, which is shown in (a) and (b). This suggests that dynamically adjusting the extraction and integration of relevant graph structural information, based on the specific question, helps the model more accurately identify key entities, ultimately contributing to improved reasoning performance.

\subsubsection{How important is Reflect?}
\begin{figure*}[h]
\includegraphics[width=\textwidth]{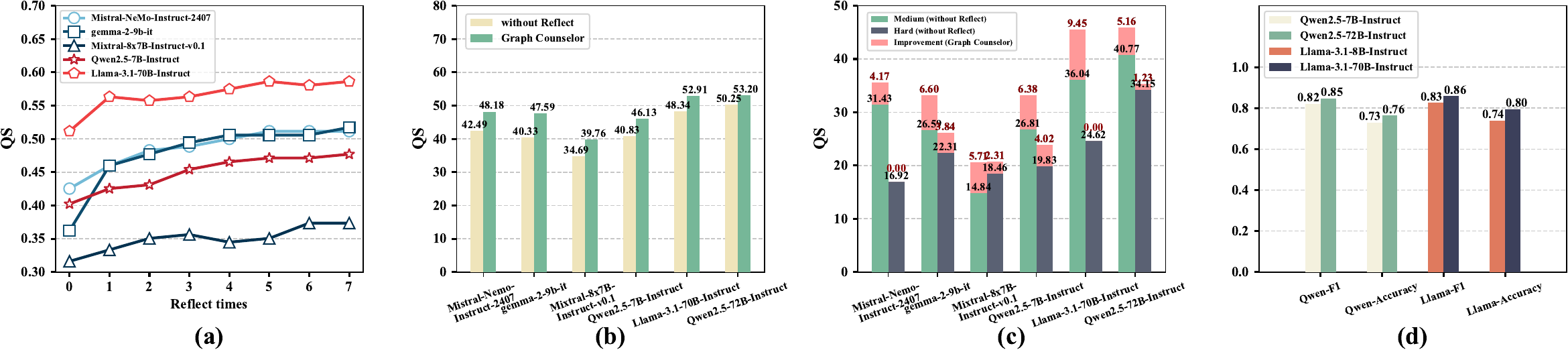}
  \caption{(a): Performance of Graph Counselor with different reflect times; (b): Results comparison between without Reflect and Graph Counselor on GRBENCH; (c): Results comparison between without Reflect and Graph Counselor on GRBENCH with different levels;
  (d): Reflect performance of models with different sizes.}
  \label{fig:reflect_performance}
  \vspace{-10pt}
\end{figure*}

\paragraph{Impact of Reflection.} To assess the role of the SR module in the Graph Counselor framework, we conduct an ablation study by removing the SR module while keeping all other experimental settings unchanged. The results are presented in Figure \ref{fig:reflect_performance} (b) and (c). Our findings show that removing the SR module leads to a performance drop of up to 7.26\% in accuracy overall, confirming its effectiveness in enhancing reasoning capabilities. Specifically, SR helps refine the model’s semantic understanding of queries while adjusting the extraction of graph structural information, thereby improving the accuracy of retrieving relevant entities based on contextual information and ultimately strengthening the model’s reasoning performance.
\paragraph{Impact of the Number of Reflection Iterations.} To determine the impact of the number of reflection iterations on the Graph Counselor framework, we conduct experiments under the same settings, testing multiple models with varying numbers of reflection iterations. The results, shown in Figure~\ref{fig:reflect_performance} (a), indicate that as the number of reflection iterations increases, the performance of Graph Counselor consistently improves. Notably, the most significant performance gain is observed at two reflection iterations, after which the improvement rate slows down. Considering the trade-off between performance gains and computational cost, we adopt two reflection iterations for all other experiments in this paper.

\paragraph{Impact of Reflection Model Size.} To investigate the impact of reflection model size on the Graph Counselor framework, we conduct an empirical analysis using different model combinations, including Qwen2.5-72B-Instruct vs. Qwen2.5-7B-Instruct and Llama-3.1-70B-Instruct vs. Llama-3.1-8B-Instruct, to evaluate their reflection capabilities. Specifically, we randomly sample 500 reasoning outputs from these models for comparative analysis, and the results are presented in Figure~\ref{fig:reflect_performance} (d). 

A key finding is that the difference in reflection performance between large and small models is not statistically significant, suggesting that model size is not the decisive factor in reflection tasks. We hypothesize three possible reasons for this observation:
\begin{itemize}
\item \textbf{Graph structure understanding is not inherently encoded in LLMs}. Reflection tasks often require extracting graph structures from textual input and analyzing reasoning validity. However, larger models do not necessarily exhibit stronger graph comprehension, as Transformer-based architectures are primarily optimized for sequential data processing rather than structured information extraction. This may explain why the performance gap between 7B and 70B models as reflection modules remains minimal; 

\item \textbf{Reflection tasks heavily rely on local logical reasoning and consistency verification.} Checking the correctness of graph structures, validating logical inference, and assessing answer accuracy are tasks that depend more on local textual coherence and logical consistency rather than extensive world knowledge. Smaller models (e.g., 7B) may already possess sufficient capabilities for these tasks, limiting the performance gain from using significantly larger models (e.g., 70B); 

\item \textbf{Reflection and self-verification abilities fall under metacognitive skills, which are not explicitly optimized in most LLM training paradigms.} Pretraining and instruction tuning of LLMs primarily focus on generating fluent and contextually coherent text rather than developing mechanisms for self-assessment and error detection. Consequently, regardless of model size, if the training data and objectives do not emphasize reflection and verification, larger models do not automatically acquire stronger self-reflection capabilities.
\end{itemize}
Therefore, considering the computational resources required for inference with 70B and 72B models, all experiments in this study utilize Qwen2.5-7B-Instruct as the reflection model.

\subsection{Trade-off between time and performance}
To further illustrate the trade-offs between time and efficiency for Graph Counselor, we report the average reasoning time (in seconds per query) required by Graph Counselor and direct reasoning using the Base Model on GRBENCH in Table \ref{tab:time}. We also include the time statistics for Graph-CoT, a baseline method that does not employ a dynamic reasoning process. Compared to these methods, our approach demonstrates the potential to achieve higher reasoning performance at a lower reasoning cost. Specifically, the results in Table \ref{tab-main overall performance} show that Graph Counselor can scale the performance of a 9B model (gemma-2-9b-it) to surpass that of Graph-CoT using a 70B model (Llama-3.1-70B-Instruct) by more than $10\%$ on the E-commerce dataset. However, the actual reasoning cost is only $13.71\%$ of that of Graph-CoT. This clearly illustrates that our method achieves higher reasoning efficiency in a relative sense, although it increases the absolute time required for reasoning compared to Graph-CoT.

\begin{table}[h]
\centering
\footnotesize
\renewcommand{\arraystretch}{1.0}
\caption{Comparison of average reasoning time (seconds per query) across different domains on GRBENCH.}
\label{tab:time}
\begin{tabular}{cccc}
\toprule
\multirow{2}{*}{Domain} & \multicolumn{3}{c}{\textbf{gemma-2-9b-it}} \\
 \cmidrule{2-4}
 & Base & Graph-CoT & Graph Counselor \\
\midrule
Academic & 0.37 & 22.30 & 54.40 \\
E-commerce & 0.47 & 24.70 & 40.30 \\
Literature & 0.40 & 14.50 & 28.25 \\
Healthcare & 0.45 & 10.00 & 32.00 \\
Legal & 0.49 & 21.00 & 44.40 \\
\midrule
\multirow{2}{*}{Domain} & \multicolumn{3}{c}{\textbf{Mistral-NeMo-2407}} \\
 \cmidrule{2-4}
 & Base & Graph-CoT & Graph Counselor \\
\midrule
Academic & 0.44 & 32.60 & 43.70 \\
E-commerce & 0.33 & 22.50 & 32.60 \\
Literature & 0.28 & 14.30 & 44.70 \\
Healthcare & 0.40 & 13.10 & 55.60 \\
Legal & 0.37 & 57.60 & 85.70 \\
\midrule
\multirow{2}{*}{Domain} & \multicolumn{3}{c}{\textbf{Llama-3-70B}}\\
 \cmidrule{2-4}
 & Base & Graph-CoT & Graph Counselor \\
\midrule
Academic & 20.65 & 269.20 & 313.30 \\
E-commerce & 13.33 & 294.00 & 390.10 \\
Literature & 7.47 & 318.80 & 489.00 \\
Healthcare & 7.61 & 320.00 & 626.70 \\
Legal & 10.13 & 240.00 & 303.30 \\
\bottomrule
\end{tabular}
\end{table}

\section{Conclusion}
In this work, we investigated the challenges encountered in the development of GraphRAG, specifically the inability to perform complex reasoning on graphs and the misalignment between graph structures and semantic information. To address these issues, we innovatively proposed Graph Counselor, a multi-round interactive iterative paradigm that systematically enables adaptive graph knowledge extraction and fully leverages the self-reflective capabilities of large models. The reasoning process of Graph Counselor can be divided into four steps: Plan, Thought, Execution, and Reflection. Subsequently, in our experiments, we tested and validated the superior performance of Graph Counselor on six backbone LLMs. Future work could focus on optimizing the efficiency and interpretability of interactive iteration mechanisms. Additionally, investigating dynamic graph updating algorithms and multimodal knowledge representation methods may further enhance reasoning generalization capabilities in open-domain scenarios.

\section*{Acknowledgements}
This work is supported by the Shanghai Municipal Science and Technology Major Project.

\section*{Limitations}
In the ablation experiments, we observed that the size of the reflection model might influence the effectiveness of Graph Counselor. However, since this phenomenon is not directly related to the core objective of this paper—enhancing the model's reasoning and comprehension capabilities—we did not perform further analysis. Our research focus consistently centered on improving the model's core abilities, and thus we did not delve into the potential relationship between reflection model size and Graph Counselor.

\bibliography{custom}

\newpage
\mbox{}
\newpage

\appendix

\onecolumn
\section{Graph Counselor Flow}
\label{sec:appendixA}
\begin{algorithm*}
\caption{Graph Counselor Flow}
\begin{algorithmic}[1]
\REQUIRE problem query $q$, graph $G$, maximum iteration times $T$, maximum reflection times $N$, Graph Knowledge Extraction $GKE$, Planning Agent $\mathrm{M}_p$, Thought Agent $\mathrm{M}_t$, Execution Agent $M_a$, Reflection Agent $\mathrm{M}_r$, Reflection model $\mathrm{M}_e$, initial context $C_0 = \emptyset$
\ENSURE final answer $y_{final}$

\STATE Initialize:
\STATE $C \gets C_0$, $y_{final} \gets \emptyset$ // Initialize context and final answer
\STATE $n \gets 0$, $correct \gets \text{False}$ // Initialize reflection count, correctness flag
\WHILE{$n \leq N$ and $\mathrm{not} \ correct $}
    \STATE $t \gets 1$ // Reset iteration step
    \WHILE{$t \leq T$}
        \STATE $P_t \gets \mathrm{M}_p(q, C)$ // Generate reasoning path
        \STATE $T_t \gets \mathrm{M}_t(q, C, P_t)$ // Clarify specific graph knowledge needed
        \STATE $A_t \gets \mathrm{M}_a(q, C, P_t, T_t)$ // Execute reasoning
        \IF{FinishCondition($A_t$)}
            \STATE $y_{final} \gets \text{Regularize}(A_t)$ // Extract answers from text based on specific rules
            \STATE $correct \gets \mathrm{M}_e(q, C, y_{final})$ // Check correctness of the result
            \STATE \textbf{break}
        \ELSE
            \STATE $O_t \gets GKE(A_t)$
            \STATE $C \gets \text{UpdateContext}(C, P_t, T_t, E_t, O_t)$ // Update context for next step
        \ENDIF
        \STATE $t \gets t + 1$ // Increment iteration step
    \ENDWHILE
    \IF{$\mathrm{not} \ correct$ and $n < N$}
        \STATE $F_n \gets M_r(C, q)$ // Generate reflective summaries based on reasoning processes
        \STATE $C \gets \text{UpdateContext}(F_n, q)$ // Update context based reflection
        \STATE $n \gets n + 1$ // Increment reflection count
    \ENDIF
\ENDWHILE
\RETURN $y_{final}$
\end{algorithmic}
\end{algorithm*}

\clearpage

\section{Additional Performance}
\renewcommand{\arraystretch}{1.45}
\label{sec:appendixB}
\begin{table*}[!h]
  \caption{\label{overall performance}
     Model performance(\%) of various methods based on Rouge-L(RL) and LlamaScore(LS)
  }
  \centering
  \fontsize{8}{11}\selectfont
  \begin{tabularx}{\linewidth}{>{\centering\arraybackslash}X>
  {\centering\arraybackslash}p{3.3cm}>{\centering\arraybackslash\columncolor{gray!10}}X>{\centering\arraybackslash}X>{\centering\arraybackslash\columncolor{gray!10}}X>{\centering\arraybackslash}X>{\centering\arraybackslash\columncolor{gray!10}}X>{\centering\arraybackslash}X>{\centering\arraybackslash\columncolor{gray!10}}X>{\centering\arraybackslash}X>{\centering\arraybackslash\columncolor{gray!10}}X>{\centering\arraybackslash}X>{\centering\arraybackslash\columncolor{gray!10}}X>{\centering\arraybackslash}X>{\centering\arraybackslash\columncolor{gray!10}}X>{\centering\arraybackslash}X>{\centering\arraybackslash\columncolor{gray!10}}X} 
    \toprule
    \multirow{2}{*}{} & \multirow{2}{*}{ \textbf{Model}} & \multicolumn{2}{c}{\textbf{Academic}} & \multicolumn{2}{c}{\textbf{E-commerce}} & \multicolumn{2}{c}{\textbf{Literature}} & \multicolumn{2}{c}{\textbf{Healthcare}} & \multicolumn{2}{c}{\textbf{Legal}}\\
    & &\textbf{RL} & \textbf{LS} &\textbf{RL} & \textbf{LS} &\textbf{RL} &\textbf{LS} &\textbf{RL} &\textbf{LS} &\textbf{RL} &\textbf{LS}\\
    \midrule
           \multirow{6}{*}{\rotatebox{90}{\textbf{Base}}} & Qwen2.5-7B-Instruct  & 9.50 & 9.53 & 10.13 & 9.00 & 6.72 & 17.08 & 7.91 & 9.26 & 26.74 & 20.00\\
           & gemma-2-9b-it  & 9.57 &  9.02 & 12.05 & 9.50 & 7.82 & 17.50 & 7.42 & 5.93 & 15.67 & 12.22 \\
           & Mistral-NeMo-Instruct-2407  & 6.34 & 5.00 & 3.37 & 3.00 & 6.33 & 11.25 & 6.14 & 5.18 & 11.56 & 8.33\\
           & Mixtral-8x7b  & 3.65 & 9.29 & 10.29 & 9.50 & 2.60 & 17.50 & 2.64 & 8.15 & 8.89 & 13.33\\
           & Llama-3.1-70B-Instruct  & 12.79 & 11.06 & 11.93 & 7.00 & 2.04 & 6.67 & 9.69 & 6.67 & 16.93 & 15.00\\
           & Qwen2.5-72B-Instruct  & 10.02 & 8.70 & 14.21 & 11.00 & 11.10 & 19.17 & 7.85 &5.93 & 29.01 & 18.89 \\
    \midrule
           \multirow{6}{*}{\rotatebox{90}{\textbf{Text RAG}}} & Qwen2.5-7B-Instruct  & 9.72 & 8.35 & 15.49 & 17.00 & 10.74 & 21.25 & 6.41 & 8.52 & 34.59 & 30.56\\
           & gemma-2-9b-it  & 9.67 &  9.21 & 19.19 & 16.50 & 13.56 & 19.17 & 4.57 & 3.70 & 30.05 & 27.78\\
           & Mistral-NeMo-Instruct-2407  & 7.22 & 6.24 & 13.33 & 11.00 & 9.68 &  14.17 & 4.33 & 3.33 & 21.73 & 18.89\\
           & Mixtral-8x7b  & 8.83 &  8.24 & 21.99 & 18.50 & 13.99 & 20.83 & 2.95 &  9.26 & 19.75 & 21.11\\
           & Llama-3.1-70B-Instruct & 14.50 & 14.47 & 20.44 &  14.50 & 14.14 & 19.58 & 7.74 & 6.67 & 28.85 & 28.89\\
           & Qwen2.5-72B-Instruct  & 11.27 & 9.18 & 24.75 &  22.00 & 15.90 & 21.67 & 8.65 & 7.78 & 37.73 & 29.44\\
    \midrule
           \multirow{6}{*}{\rotatebox{90}{\textbf{GraphRAG (1-hop)}}} & Qwen2.5-7B-Instruct  & 26.71 & 26.47 & 24.70 & 23.00 & 15.15 & 23.75 & 11.17 & 13.33 & 38.58 & 34.44\\
           & gemma-2-9b-it & 30.70 &  28.44 & 27.10 & 23.00 & 21.00 & 25.00 & 17.48 & 14.07 & 26.66 & 23.33\\
           & Mistral-NeMo-Instruct-2407  & 20.08 & 18.47 & 15.62 & 13.00 & 15.20 & 18.33 & 11.32 & 10.37 & 26.80 & 25.56\\
           & Mixtral-8x7b  & 30.47 &  30.12 & 30.99 & 25.50 & 18.87 & 25.83 & 8.58 &  13.70 & 25.21 & 27.78\\
           & Llama-3.1-70B-Instruct  & 32.96 & 32.24 & 29.98 & 25.00 & 24.47 & 31.67 & 21.19 & 17.04 & 41.33 &  36.11\\
           & Qwen2.5-72B-Instruct  & 37.68 & 33.53 & 32.11 & 28.00 & 23.65 & 31.25 & 19.69 & 18.52 & 41.14 & 35.00\\
    \midrule
           \multirow{6}{*}{\rotatebox{90}{\textbf{GraphRAG (2-hop)}}} & Qwen2.5-7B-Instruct  & 27.98 & 26.24 & 20.87 & 20.00 & 16.32 & 26.67 & 7.34 & 9.26 & 40.62 & 39.44\\
           & gemma-2-9b-it & 31.36 & 27.74 & 21.77 & 19.00 & 21.34 & 25.00 & 2.82 &  2.59 & 32.08 & 28.33\\
           & Mistral-NeMo-Instruct-2407  & 14.40 & 13.06 & 15.49 & 13.50 & 14.41 & 19.17 & 4.24 & 2.96 & 23.50 & 24.44\\
           & Mixtral-8x7b  & 24.29 &  24.50 & 25.23 & 21.00 & 19.45 & 28.33 & 4.77 &  8.15 & 27.18 & 26.11\\
           & Llama-3.1-70B-Instruct  & 33.09 & 32.94 & 26.36 & 25.00 & 23.46 & 30.00 & 11.46 & 8.52 & 42.52 & 37.22\\
           & Qwen2.5-72B-Instruct  & 36.74 & 31.88 & 30.39 &  25.50 & 25.60 & 30.42 & 8.41 & 7.04 & 43.58 & 38.33\\
    \midrule
           \multirow{6}{*}{\rotatebox{90}{\textbf{Graph-CoT}}} & Qwen2.5-7B-Instruct  & 38.27 & 40.82 & 39.77 &  40.50 & 37.08 & 45.83 & 34.88 & 37.04 & 31.63 &  36.67\\
           & gemma-2-9b-it  & 41.51 & 40.89 & 37.10 & 40.50 & 41.25 & 45.83 & 29.50 & 33.70 & 28.12 & 32.46\\
           & Mistral-NeMo-Instruct-2407  & 32.26 & 34.50 & 30.33 & 39.50 & 24.67 &  36.67 & 27.26 & 32.59 & 29.09 & 36.11\\
           & Mixtral-8x7b  & 31.78 &  31.53 & 29.57 & 33.00 & 35.61 & 41.67 & 27.26 &  25.56 & 17.93 & 26.11\\
           & Llama-3.1-70B-Instruct  & 47.64 & 50.47 & 31.21 &  36.00 & 42.06 & 45.83 & 43.70 & 48.52 & 41.60 & 43.89\\
           & Qwen2.5-72B-Instruct  & 51.76 & 57.76 & 25.34 &  33.00 & 38.67 & 49.58 & 45.26 & 50.00 & 41.61 & 42.22\\
    \midrule
           \multirow{6}{*}{\rotatebox{90}{\textbf{Graph Counselor}}} & Qwen2.5-7B-Instruct  & 47.80 & 46.35 & 46.73 & 43.50 & 47.10 & 51.67 & 42.18 & 42.59 & 48.36 & 49.44\\
           & gemma-2-9b-it  & 55.58 & 51.69 & 49.02 & 49.50 & 55.41 & 58.75 & 42.21 & 43.70 & 35.74 & 40.00\\
           & Mistral-NeMo-Instruct-2407  & 54.15 & 51.21 & 44.46 & 44.00 & 47.71 & 53.75 & 43.87 & 43.33 & 53.35 & 54.44\\
           & Mixtral-8x7b  & 44.72 & 41.46 & 44.71 & 44.00 & 44.07 & 46.25 & 34.98 & 38.15 & 25.75 & 32.22\\
           & Llama-3.1-70B-Instruct  & 60.11 & 60.67 & 48.33 & 48.50 & 56.31 & 61.67 & 48.90 & 48.89 & 53.84 & 54.44\\
           & Qwen2.5-72B-Instruct  & 61.53 & 63.04 & 37.21 & 45.00 & 48.55 & 56.67 & 48.06 & 47.78 & 57.79 & 58.89\\
    \bottomrule   
  \end{tabularx}
\end{table*}

\twocolumn
\section{Case Studies of Graph Counselor}
To verify the effectiveness of Graph Counselor in self-reflection and error correction, we provide two erroneous examples (the \textbf{Amazon} and \textbf{Biomedical} as shown below) and the corrected examples after the intermediate reflection process. The results indicate that the reflection process can indeed effectively identify the key points where problems arise and improve the reasoning process based on these key points to arrive at the correct answers.
\begin{tcolorbox}[colback=white!10!white, colframe=blue!30!black, title=Wrong Case (Amazon), boxrule=1pt, fontupper={\fontsize{7pt}{9pt}\selectfont}]
\justifying
\noindent
\textbf{Question}: What is the quantity of items sharing the same purchased-together items as item Nokia CC-3068 Shell for Lumia 520 - Retail Packaging - White? Please answer by providing node main feature (e.g., names) rather than node IDs. \\
\textbf{Plan 1}: The question is asking for the count of items that share the same bought\_together\_item neighbors as the specified item, excluding the original item. We need to locate the specified item, list its bought\_together\_item neighbors, and then find other items that share these neighbors. Finally, we need to exclude the original item and count the remaining items.\\
\textbf{Thought 1}: We need to find the node in the graph that represents the specified item.\\
\textbf{Action 1}: Retrieve[Nokia CC-3068 Shell for Lumia 520 - Retail Packaging - White]\\
\textbf{Observation 1}: The ID of this retrieval target node is B00BRPTT9K. \\
\textbf{Plan 2}: We have the node representing the specified item. Next, we need to list its bought\_together\_item neighbors.\\
\textbf{Thought 2}: We need to retrieve the bought\_together\_item neighbors of the specified item.\\
\textbf{Action 2}: Neighbour[B00BRPTT9K, bought\_together\_item]\\
\textbf{Observation 2}: The bought\_together\_item neighbors of B00BRPTT9K are: ['B00E45043A']. \\
\textbf{Plan 3}: We have the bought\_together\_item neighbors of the specified item. Next, we need to find other items that share these neighbors.\\
\textbf{Thought 3}: We need to find items that share the same bought\_together\_item neighbors as the specified item.\\
\textbf{Action 3}: Neighbour[B00E45043A, bought\_together\_item]\\
\textbf{Observation 3}: The bought\_together\_item neighbors of B00E45043A are: [...](49 nodes including B00BRPTT9K) \\
\textbf{Plan 4}: We have the list of items that share the same bought\_together\_item neighbors as the specified item. Next, we need to filter out the original item and count the remaining items.\\
\textbf{Thought 4}: We need to filter out the original item (B00BRPTT9K) from the list and count the remaining items.\\
\textbf{Action 4}: Feature[Retrieve[B00BRPTT9K], title]\\
\textbf{Observation 4}: The ID of this retrieval target node is brand\_70532. The node or feature name does not exist in the graph. This might because your given feature name is not correct. Please modify it.\\
\textbf{Plan 5}: It seems there was an error in retrieving the title of the original item. We will directly filter out the original item from the list and count the remaining items.\\
\textbf{Thought 5}: We need to filter out the original item (B00BRPTT9K) from the list and count the remaining items.\\
\textbf{Action 5}: Finish[49]\\
\textbf{Observation 5}: Answer is 49.\\
\textbf{Correct answer}: 48
\end{tcolorbox}

\begin{tcolorbox}[colback=white!10!white, colframe=blue!30!black, title=Reflection (Amazon), boxrule=1pt, fontupper={\fontsize{7pt}{9pt}\selectfont}]
\justifying
\noindent
\raggedright
You have attempted to answer following question before and failed. The following reflection(s) give a plan to avoid failing to answer the question in the same way you did previously. Use them to improve your strategy of correctly answering the given question.
Reflections:\\
- Recap of the Trial\\
    - Question: What is the quantity of items sharing the same purchased-together items as item Nokia CC-3068 Shell for Lumia 520 - Retail Packaging - White? Please answer by providing node main feature (e.g., names) rather than node IDs.\\
    - Graph Information Used:\\
  - Retrieve[Nokia CC-3068 Shell for Lumia 520 - Retail Packaging - White]: To locate the node representing the specified item.\\
  - Neighbour[B00BRPTT9K, bought\_together\_item]: To list the bought\_together\_item neighbors of the specified item.\\
  - Neighbour[B00E45043A, bought\_together\_item]: To list the items that share the same bought\_together\_item neighbors.\\
- Outcome: The question was not successfully answered due to guessing the wrong answer with Finish[].\\
- Guided Reflection\\
  - Understanding the Question\\
    - Core Goal: Provide the count of items that share the same bought\_together\_item neighbors as the specified item, excluding the original item.\\
    - Missed Information: Misunderstood the requirement to filter out the original item and count the remaining items.\\
    - Potential Misunderstandings: Assumed the answer was the total count of items sharing the same bought\_together\_item neighbors, including the original item.\\
  - Analysis of Selected Graph Information\\
    - Relevance: Retrieve: Correct step to locate the item node. Neighbour: Appropriate to list the bought\_together\_item neighbors and items sharing these neighbors.\\
    - Missed Insights: Overlooked the need to filter out the original item before counting the remaining items.\\
    - Redundancies: The process was unnecessarily complex, leading to potential errors in counting.\\

  - Align the Problem Understanding and Graph Function Information\\
    - Understanding and Selection: Misaligned the use of Neighbour, leading to an incorrect count.\\
    - Inconsistencies: Misunderstood the requirement to exclude the original item, leading to an incorrect answer.\\
    - Adjustments: Focus on using Neighbour to list the items sharing the same bought\_together\_item neighbors and then filter out the original item before counting.\\
- Improved Strategy\\
  - Updated Understanding of the Problem:
  The question requires the count of items that share the same bought\_together\_item neighbors as the specified item, excluding the original item.\\
  - Revised Graph Selection:\\
    - Graph Selection\\
    Use Retrieve[Nokia CC-3068 Shell for Lumia 520 - Retail Packaging - White] to locate the target node.\\
    Use Neighbour[B00BRPTT9K, bought\_together\_item] to list the bought\_together\_item neighbors of the specified item.\\
    Use Neighbour for each neighbor to list items sharing the same bought\_together\_item neighbors.\\
    Filter out the original item and count the remaining items.\\
    - Combine Functions suggestions\\
    You can combine Retrieve with Neighbour to streamline the process:\\
    Neighbour[Retrieve[Nokia CC-3068 Shell for Lumia 520 - Retail Packaging - White], bought\_together\_item]\\
    - Further combine Neighbour to list items sharing the same neighbors:\\
    Neighbour[Neighbour[Retrieve[Nokia CC-3068 Shell for Lumia 520 - Retail Packaging - White], bought\_together\_item], bought\_together\_item]\\
  - Avoiding Past Issues:\\
  Focus on the core requirement (providing the count of items sharing the same bought\_together\_item neighbors, excluding the original item).\\
  Ensure the original item is filtered out before counting.\\
  Use concise and relevant graph operations to align with the problem's objective. \\
  (END OF REFLECTION)
\end{tcolorbox}

\begin{tcolorbox}[colback=white!10!white, colframe=blue!30!black, title=True Case (Amazon), boxrule=1pt, fontupper={\fontsize{7pt}{9pt}\selectfont}]
\justifying
\noindent
\raggedright
\textbf{Question}: What is the quantity of items sharing the same purchased-together items as item Nokia CC-3068 Shell for Lumia 520 - Retail Packaging - White? Please answer by providing node main feature (e.g., names) rather than node IDs. \\
\textbf{Plan 1}: The question is asking for the count of items that share the same bought\_together\_item neighbors as the specified item, excluding the original item. We need to locate the specified item, list its bought\_together\_item neighbors, and then find other items that share these neighbors. Finally, we need to exclude the original item and count the remaining items.\\
\textbf{Thought 1}: We need to find the node in the graph that represents the specified item.\\
\textbf{Action 1}: Retrieve[Nokia CC-3068 Shell for Lumia 520 - Retail Packaging - White]\\
\textbf{Observation 1}: The ID of this retrieval target node is B00BRPTT9K. \\
\textbf{Plan 2}: We have the node representing the specified item. Next, we need to list its bought\_together\_item neighbors.\\
\textbf{Thought 2}: We need to retrieve the bought\_together\_item neighbors of the specified item.\\
\textbf{Action 2}: Neighbour[B00BRPTT9K, bought\_together\_item]\\
\textbf{Observation 2}: The bought\_together\_item neighbors of B00BRPTT9K are: ['B00E45043A']. \\
\textbf{Plan 3}: We have the bought\_together\_item neighbors of the specified item. Next, we need to find other items that share these neighbors.\\
\textbf{Thought 3}: We need to find items that share the same bought\_together\_item neighbors as the specified item.\\
\textbf{Action 3}: Neighbour[B00E45043A, bought\_together\_item]\\
\textbf{Observation 3}: The bought\_together\_item neighbors of B00E45043A are: [...](49 nodes including B00BRPTT9K)\\ 
\textbf{Plan 4}: We have the list of items that share the same bought\_together\_item neighbors as the specified item. Next, we need to filter out the original item and count the remaining items.\\
\textbf{Thought 4}: We need to filter out the original item (B00BRPTT9K) from the list and count the remaining items.\\
\textbf{Action 4}: Finish[48]\\
\textbf{Observation 4}: Answer is 48\\
\textbf{Correct answer}: 48
\end{tcolorbox}

\begin{tcolorbox}[colback=white!10!white, colframe=blue!30!black, title=Wrong Case (Biomedical), boxrule=1pt, fontupper={\fontsize{7pt}{9pt}\selectfont}]
\justifying
\noindent
\raggedright
\textbf{Question}: What illness situated in ear can be treated by Fluocinolone Acetonide? Please answer by providing node main feature (e.g., names) rather than node IDs. \\
\textbf{Plan 1}: The question is asking for a disease localized in the ear that can be treated by Fluocinolone Acetonide. We need to find the node representing the compound and then identify the diseases it treats. Next, we need to check which of these diseases are localized in the ear.\\
\textbf{Thought 1}: We need to find the node representing Fluocinolone Acetonide in the graph.\\
\textbf{Action 1}: Retrieve[Fluocinolone Acetonide]\\
\textbf{Observation 1}: The ID of this retrieval target node is DB00591. \\
\textbf{Plan 2}: The question is asking for a disease localized in the ear that can be treated by Fluocinolone Acetonide. We have the node representing the compound. Next, we need to find the diseases that this compound treats.\\
\textbf{Thought 2}: We need to find the diseases that Fluocinolone Acetonide treats.\\
\textbf{Action 2}: Neighbour[DB00591, Compound-treats-Disease]\\
\textbf{Observation 2}: The Compound-treats-Disease neighbors of DB00591 are: ['DOID:3310', 'DOID:8893']. \\
\textbf{Plan 3}: The question is asking for a disease localized in the ear that can be treated by Fluocinolone Acetonide. We have the diseases that this compound treats. Next, we need to check which of these diseases are localized in the ear.\\
\textbf{Thought 3}: We need to check the anatomical locations of the diseases 'DOID:3310' and 'DOID:8893'.\\
\textbf{Action 3}: Neighbour[DOID:3310, Disease-localizes-Anatomy], Neighbour[DOID:8893, Disease-localizes-Anatomy]\\
\end{tcolorbox}

\begin{tcolorbox}[colback=white!10!white, colframe=blue!30!black, title=, boxrule=1pt, fontupper={\fontsize{7pt}{9pt}\selectfont}]
\justifying
\noindent
\textbf{Observation 3}: The Disease-localizes-Anatomy neighbors of DOID:3310 are: [...](17 nodes) The Disease-localizes-Anatomy neighbors of DOID:8893 are: [...](24 nodes)\\
\textbf{Plan 4}: The question is asking for a disease localized in the ear that can be treated by Fluocinolone Acetonide. We have the anatomical locations of the diseases. Next, we need to identify which of these locations correspond to the ear.\\
\textbf{Thought 4}: We need to get the names of the anatomical locations to identify which ones correspond to the ear.\\
\textbf{Action 4}: Feature[UBERON:0001690, name], Feature[UBERON:0001691, name], ...\\
\textbf{Observation 4}: The name feature of UBERON:0001690 are: ear. The name feature of UBERON:0001691 are: external ear. ...\\ 
\textbf{Plan 5}: The question is asking for a disease localized in the ear that can be treated by Fluocinolone Acetonide. We have the names of the anatomical locations. Next, we need to identify which of these locations correspond to the ear and then get the names of the diseases associated with these locations.\\
\textbf{Thought 5}: We need to identify the diseases associated with the ear and external ear.\\
\textbf{Action 5}: Feature[DOID:3310, name], Feature[DOID:8893, name]\\
\textbf{Observation 5}: The name feature of DOID:3310 are: atopic dermatitis. The name feature of DOID:8893 are: psoriasis. \\
\textbf{Plan 6}: The question is asking for a disease localized in the ear that can be treated by Fluocinolone Acetonide. We have the names of the diseases. Next, we need to verify which of these diseases are localized in the ear.\\
\textbf{Thought 6}: We need to verify which of the diseases 'atopic dermatitis' and 'psoriasis' are localized in the ear.\\
\textbf{Action 6}: Feature[DOID:3310, name], Feature[DOID:8893, name]\\
\textbf{Observation 6}: The name feature of DOID:3310 are: atopic dermatitis. The name feature of DOID:8893 are: psoriasis. \\
\textbf{Plan 7}: The question is asking for a disease localized in the ear that can be treated by Fluocinolone Acetonide. We have the names of the diseases. We need to verify which of these diseases are localized in the ear.\\
\textbf{Thought 7}: We need to verify the anatomical locations of 'atopic dermatitis' and 'psoriasis' again to ensure they are localized in the ear.\\
\textbf{Action 7}: Neighbour[DOID:3310, Disease-localizes-Anatomy], Neighbour[DOID:8893, Disease-localizes-Anatomy]\\
\textbf{Observation 7}: The Disease-localizes-Anatomy neighbors of DOID:3310 are: [...](17 nodes) The Disease-localizes-Anatomy neighbors of DOID:8893 are: [...](24 nodes)
\textbf{Plan 8}: The question is asking for a disease localized in the ear that can be treated by Fluocinolone Acetonide. We have the anatomical locations of the diseases. Next, we need to identify which of these locations correspond to the ear.\\
\textbf{Thought 8}: We need to get the names of the anatomical locations to identify which ones correspond to the ear.\\
\textbf{Action 8}: Feature[UBERON:0001690, name], Feature[UBERON:0001691, name]\\
\textbf{Observation 8}: The name feature of UBERON:0001690 are: ear. The name feature of UBERON:0001691 are: external ear. \\
\textbf{Plan 9}: The question is asking for a disease localized in the ear that can be treated by Fluocinolone Acetonide. We have the names of the anatomical locations. Next, we need to verify which of these diseases are localized in the ear.\\
\textbf{Thought 9}: We need to verify which of the diseases 'atopic dermatitis' and 'psoriasis' are localized in the ear.\\
\textbf{Action 9}: Feature[DOID:3310, name], Feature[DOID:8893, name]\\
\textbf{Observation 9}: The name feature of DOID:3310 are: atopic dermatitis. The name feature of DOID:8893 are: psoriasis. \\
\textbf{Plan 10}: The question is asking for a disease localized in the ear that can be treated by Fluocinolone Acetonide. We have the names of the diseases and their anatomical locations. Next, we need to identify which disease is localized in the ear.\\
\textbf{Thought 10}: We need to identify which of the diseases 'atopic dermatitis' and 'psoriasis' are localized in the ear.\\
\textbf{Action 10}: Feature[DOID:3310, name]\\
Observation 10: The name feature of DOID:3310 are: atopic dermatitis.\\
\textbf{Correct answer}: atopic dermatitis\\
\end{tcolorbox}

\begin{tcolorbox}[colback=white!10!white, colframe=blue!30!black, title=Reflection (Biomedical), boxrule=1pt, fontupper={\fontsize{7pt}{9pt}\selectfont}]
\justifying
\noindent
\raggedright

You have attempted to answer following question before and failed. The following reflection(s) give a plan to avoid failing to answer the question in the same way you did previously. Use them to improve your strategy of correctly answering the given question.
Reflections:\\
- - Recap of the Trial\\
  - Question: What illness situated in ear can be treated by Fluocinolone Acetonide? Please answer by providing node main feature (e.g., names) rather than node IDs.\\
  - Graph Information Used:\\
    Retrieve[Fluocinolone Acetonide]: To locate the node representing the compound.\\
    Neighbour[DB00591, Compound-treats-Disease]: To find the diseases treated by the compound.\\
    Neighbour[DOID:3310, Disease-localizes-Anatomy], Neighbour[DOID:8893, Disease-localizes-Anatomy]: To find the anatomical locations of the diseases.\\
    Feature[UBERON:0001690, name], Feature[UBERON:0001691, name], Feature[UBERON:0001037, name]: To get the names of the anatomical locations.\\
    Feature[DOID:3310, name]: To get the name of the disease.\\
  - Outcome: The question was successfully answered by identifying atopic dermatitis as the disease localized in the ear that can be treated by Fluocinolone Acetonide.\\
- Guided Reflection\\
  - Understanding the Question\\
    - Core Goal: Identify a disease localized in the ear that can be treated by Fluocinolone Acetonide.\\
    - Missed Information: None.\\
    - Potential Misunderstandings: None.\\
  - Analysis of Selected Graph Information\\
    - Relevance: All selected graph information was relevant to answering the question.\\
    - Missed Insights: None.\\
    - Redundancies: Some redundant steps were taken to verify the name of the disease multiple times.\\
  - Align the Problem Understanding and Graph Function Information\\
    - Understanding and Selection: The steps taken were aligned with the problem understanding.\\
    - Inconsistencies: None.\\
    - Adjustments: Streamline the process by reducing redundant checks.\\
- Improved Strategy\\
  - Updated Understanding of the Problem:
    The question requires identifying a disease localized in the ear that can be treated by Fluocinolone Acetonide.\\
  - Revised Graph Selection:\\
    - Graph Selection\\
      - Use Retrieve[Fluocinolone Acetonide] to locate the compound node.\\
      - Use Neighbour[DB00591, Compound-treats-Disease] to find the diseases treated by the compound.\\
      - Use Neighbour on each disease node to find their anatomical locations.\\
      - Use Feature to get the names of the anatomical locations and diseases.\\
    - Combine Functions Suggestions\\
      - Combine Neighbour and Feature to streamline the process:\\
        - Feature[Neighbour[DB00591, Compound-treats-Disease], name] to get the names of the diseases.\\
        - Feature[Neighbour[DOID:3310, Disease-localizes-Anatomy], name] to get the names of the anatomical locations.\\
  - Avoiding Past Issues:\\
    - Ensure that each step is necessary and avoid redundant checks.\\
    - Verify the anatomical location and disease names in a single step where possible.\\
    - Provide the final answer based on the verified information.\\
(End of Reflection)
\end{tcolorbox}

\begin{tcolorbox}[colback=white!10!white, colframe=blue!30!black, title=True Case (Biomedical), boxrule=1pt, fontupper={\fontsize{8pt}{10pt}\selectfont}]
\justifying
\noindent
\textbf{Question}: What illness situated in ear can be treated by Fluocinolone Acetonide? Please answer by providing node main feature (e.g., names) rather than node IDs. \\
\textbf{Plan 1}: The question is asking for a disease localized in the ear that can be treated by Fluocinolone Acetonide. We need to find the node representing the compound and then identify the diseases it treats. Next, we need to check which of these diseases are localized in the ear.\\
\textbf{Thought 1}: We need to find the node representing Fluocinolone Acetonide in the graph.\\
\textbf{Action 1}: Retrieve[Fluocinolone Acetonide]\\
\textbf{Observation 1}: The ID of this retrieval target node is DB00591. \\
\textbf{Plan 2}: The question is asking for a disease localized in the ear that can be treated by Fluocinolone Acetonide. We have the node representing the compound. Next, we need to find the diseases that this compound treats.\\
\textbf{Thought 2}: We need to find the diseases that Fluocinolone Acetonide treats.\\
\textbf{Action 2}: Neighbour[DB00591, Compound-treats-Disease]\\
\textbf{Observation 2}: The Compound-treats-Disease neighbors of DB00591 are: ['DOID:3310', 'DOID:8893']. \\
\textbf{Plan 3}: The question is asking for a disease localized in the ear that can be treated by Fluocinolone Acetonide. We have the diseases that this compound treats. Next, we need to check which of these diseases are localized in the ear.\\
\textbf{Thought 3}: We need to check the anatomical locations of the diseases 'DOID:3310' and 'DOID:8893'.\\
\textbf{Action 3}: Neighbour[DOID:3310, Disease-localizes-Anatomy], Neighbour[DOID:8893, Disease-localizes-Anatomy]\\
\textbf{Observation 3}: The Disease-localizes-Anatomy neighbors of DOID:3310 are: [...](17 nodes) The Disease-localizes-Anatomy neighbors of DOID:8893 are: [...](24 nodes) \\
\textbf{Plan 4}: The question is asking for a disease localized in the ear that can be treated by Fluocinolone Acetonide. We have the anatomical locations of the diseases. Next, we need to identify which of these locations correspond to the ear.\\
\textbf{Thought 4}: We need to get the names of the anatomical locations to identify which ones correspond to the ear.\\
\textbf{Action 4}: Feature[UBERON:0001690, name], Feature[UBERON:0001691, name], ... //
\textbf{Observation 4}: The name feature of UBERON:0001690 are: ear. The name feature of UBERON:0001691 are: external ear. ... \\
\textbf{Plan 5}: The question is asking for a disease localized in the ear that can be treated by Fluocinolone Acetonide. We have the names of the anatomical locations. Next, we need to identify which of these locations correspond to the ear and then get the names of the diseases associated with these locations.\\
\textbf{Thought 5}: We need to identify the diseases associated with the ear.\\
\textbf{Action 5}: Feature[DOID:3310, name]\\
\textbf{Observation 5}: The name feature of DOID:3310 are: atopic dermatitis. \\
\textbf{Plan 6}: The question is asking for a disease localized in the ear that can be treated by Fluocinolone Acetonide. We have the answer.\\
\textbf{Thought 6}: We have the answer: atopic dermatitis.\\
\textbf{Action 6}: Finish[atopic dermatitis]\\
\textbf{Observation 6}: Answer is atopic dermatitis\\
\textbf{Correct answer}: atopic dermatitis
\end{tcolorbox}

\begin{tcolorbox}[colback=white!10!white, colframe=blue!30!black, title=Reflection Process, boxrule=1pt, fontupper={\fontsize{7pt}{9pt}\selectfont}]
\justifying
\noindent
\textbf{- Recap of the Trial}
\begin{adjustwidth}{6pt}{0pt}
 \textbf{- Question}: Who can be defined as the closest collaborator of author Alice B. Cooper, according to the number of  joint works? Please answer by providing node main feature (e.g., names) rather than node IDs.\\
 \textbf{- Graph Information Used}:
\end{adjustwidth}
\begin{adjustwidth}{11pt}{0pt}
    Retrieve[Alice B. Cooper]: To find the node representing Alice B. Cooper.\\
    Degree[2676686768, author]: To calculate the degree of "author" neighbors.\\
    Feature[Neighbour[2676686768, author], name]: To retrieve the names of all "author" neighbors.
\end{adjustwidth}
\begin{adjustwidth}{6pt}{0pt}
\textbf{- Outcome}: The question was not successfully answered due to incorrect assumptions about the graph structure and the failure of neighbor-related functions.
\end{adjustwidth}
\textbf{- Guided Reflection}
\begin{adjustwidth}{6pt}{0pt}
 - \textbf{Understanding the Question}
\end{adjustwidth}
\begin{adjustwidth}{11pt}{0pt}
   \textit{- Core Goal}: Identify the closest collaborator of Alice B. Cooper by evaluating the number of joint works, which requires correctly analyzing the relationships between nodes in the graph.\\
   \textit{- Missed Information}: Misinterpreted or overlooked the correct relationship or neighbor type that represents "collaborators."\\
   \textit{- Potential Misunderstandings}: Misinterpreted the graph's structure, assuming direct "author" neighbors represent collaborators. The collaboration is mediated by shared "paper" nodes.
\end{adjustwidth}
\textbf{- Analysis of Selected Graph Information}
\begin{adjustwidth}{11pt}{0pt}
 \textit{- Relevance}: Retrieve[Alice B. Cooper]: Correct step to locate the node representing the author. Degree[2676686768, author]:Misapplied, as direct "author" neighbors do not exist. Feature[Neighbour[2676686768, author], name]: Irrelevant, as there are no direct "author" neighbors to retrieve.\\
 \textit{- Missed Insights}: Did not consider exploring the graph to confirm how collaborators are connected (e.g., through shared papers). Overlooked the use of Feature to extract details about "paper" connections.\\
 \textit{- Redundancies}: Repeated attempts to use "author" as the neighbor type without validating its relevance were redundant.
\end{adjustwidth}
\begin{adjustwidth}{6pt}{0pt}
 \textbf{- Align the Problem Understanding and Graph Function Information}
\end{adjustwidth}
\begin{adjustwidth}{11pt}{0pt}
 \textit{- Understanding and Selection}: The misunderstanding of the question's requirements led to inappropriate graph function calls.\\
 \textit{- Inconsistencies}: Assumed collaborators were directly connected as "author" neighbors, which was inconsistent with the graph's definition.\\
 \textit{- Adjustments}: Start by understanding the graph's structure and how nodes are linked (e.g., "author" nodes are linked through "paper" nodes). Verify the relationship structure of the graph first using exploratory functions or metadata queries.
\end{adjustwidth}
\textbf{- Improved Strategy}
\begin{adjustwidth}{6pt}{0pt}
 \textbf{- Updated Understanding of the Problem}:
\end{adjustwidth}
\begin{adjustwidth}{11pt}{0pt}
    Collaborators of Alice B. Cooper are determined by shared "paper" nodes. To find the closest collaborator, it is necessary to identify the "paper" neighbors of Alice B. Cooper, then determine the authors with the highest co-authorship count.
\end{adjustwidth}
\begin{adjustwidth}{6pt}{0pt}
  \textbf{- Revised Graph Selection}:
\end{adjustwidth}
\begin{adjustwidth}{11pt}{0pt}
    \textit{- Graph Selection}\\
    Use Retrieve[Alice B. Cooper] to find Alice B. Cooper's node.\\
    Use Neighbour with the type "paper" to list all "paper" neighbors of Alice B. Cooper.

    For each "paper" node, check its "author" neighbors using Neighbour.\\
    Count the occurrences of each author node in the "paper" neighbors list to find the collaborator with the highest joint works.\\
    Use Feature to retrieve the name of the closest collaborator.\\
 \textit{- Combine Functions suggestions}\\
    You can combine Retrieve with Neighbour to list all papers authored by Alice B. Cooper: Neighbour[Retrieve[Alice B. Cooper], paper]
\end{adjustwidth}
\begin{adjustwidth}{6pt}{0pt}
 \textbf{- Avoiding Past Issues}: 
\end{adjustwidth}
\begin{adjustwidth}{11pt}{0pt}
   Confirm graph structure and neighbor relationships before performing specific queries.\\
   Avoid redundant actions by verifying assumptions at each step.\\
   Use appropriate graph functions (e.g., Neighbour and Feature) to extract detailed relationship data.
\end{adjustwidth}
    (END OF REFLECTION)
\end{tcolorbox}

\newpage
\section{Prompts in Graph Counselor}
\label{sec:appendixD}
The prompts designed to instruct Large Language Models (LLMs) for the Graph Counselor can be categorized into three distinct types: model reasoning prompts, model reflection prompts, and model evaluation prompts.
\subsection{Model Reasoning Prompts}
\label{sec:appendixD1}

\begin{tcolorbox}[colback=white!10!white, colframe=blue!30!black, title=Reasoning Prompts, boxrule=1pt]
\justifying
\noindent
Solve a question answering task with interleaving Thought, Interaction with Graph, Feedback from Graph steps. In Plan step, you can think about what the question is asking and plan how to do to get the answer. In Thought step, you can think about what further information is needed, and In Interaction step, you can get feedback from graphs with four functions: \\
(1) Retrieve[keyword], which retrieves the related node from the graph according to the corresponding query.\\
(2) Feature[Node, feature], which returns the detailed attribute information of Node regarding the given "feature" key.\\
(3) Degree[Node, neighbor\_type], which calculates the number of "neighbor\_type" neighbors of the node Node in the graph.\\
(4) Neighbour[Node, neighbor\_type], which lists the "neighbor\_type" neighbours of the node Node in the graph and returns them.\\
Besides, you can use compound function, such as Feature[Retrieve [keyword], feature], which returns the detailed attribute information of Retrieve[keyword] regarding the given "feature" key.\\
Here are some examples:\\
\{\textit{examples}\}\\
(END OF EXAMPLES)\\
\{\textit{reflections}\}\\
When last Observation has been given or there is no Plan, you should provide next only one Plan based on the question. When last Plan has been given, you should provide next only one Thought. When last Thought has been given, you should provide next only one Action.\\
\mbox{}\\
\mbox{}\\
\end{tcolorbox}

\begin{tcolorbox}[colback=white!10!white, colframe=blue!30!black, title=, boxrule=1pt]
\justifying
\noindent
When you think it's time to finish, use Finish to end the process. Don't make Observation.\\
Definition of the graph: \\\{\textit{graph\_definition}\}\\
Question: \{\textit{question}\} Please answer by providing node main feature (e.g., names) rather than node IDs.
\end{tcolorbox}

\subsection{Model Reflection Prompts}
\label{sec:appendixD2}

\begin{tcolorbox}[colback=white!10!white, colframe=blue!30!black, title=Reflection Prompts, boxrule=1pt]
\justifying
\noindent
You are an advanced reasoning agent that can improve based on self reflection. Reflect on your prior reasoning trial, and find areas for improvement to enhance your performance in answering the question next time. Please write the Reflections, including the content for Recap of the Trial, Guided Reflection, based on the guidance provided in Graph Function Background, Previous Trial Details, Recap of the Trial, and Guided Reflection.\\
Graph Function Background\\
- Definition of the graph: \\ \{\textit{graph}\_\textit{definition}\} \\
- You were provided with the following functions to interact with the graph:\\
  - Retrieve[keyword]: Finds the related node based on the query keyword.\\
  - Feature[Node, feature]: Retrieves detailed attribute information for the specified node and feature key.\\
  - Degree[Node, neighbor\_type]: Calculates the number of neighbors of the specified type for the given node.\\
  - Neighbour[Node, neighbor\_type]: Lists the neighbors of the specified type for the given node.\\
  Besides, you can use compound function, such as Feature[Retrieve [keyword], feature], which returns the detailed attribute information of Retrieve[keyword] regarding the given "feature" key.\\
\textbf{- Recap of the Trial}
\begin{adjustwidth}{6pt}{0pt}
    \textbf{- Question}: [Insert the problem description here]\\
    \textbf{- Graph Information Used}: [List the graph structural information selected in the trial]
\end{adjustwidth}
\end{tcolorbox}

\begin{tcolorbox}[colback=white!10!white, colframe=blue!30!black, title=, boxrule=1pt]
\justifying
\noindent
\begin{adjustwidth}{6pt}{0pt}
\textbf{- Outcome}: Identify whether the attempt failed due to incorrect answers, exceeded steps (10 steps), or reasoning length limits.
\end{adjustwidth}
\textbf{- Guided Reflection}
\begin{adjustwidth}{6pt}{0pt}
\textbf{- Understanding the Question}
\end{adjustwidth}
\begin{adjustwidth}{11pt}{0pt}
 \textit{- Core Goal}: What is the main objective of this question?\\
 \textit{- Missed Information}: Could you have overlooked any critical details?\\
\textit{- Potential Misunderstandings}: Were there any misinterpretations in your approach? If so, list and correct them.
\end{adjustwidth}
\begin{adjustwidth}{6pt}{0pt}
\textbf{- Analysis of Selected Graph Information}
\end{adjustwidth}
\begin{adjustwidth}{11pt}{0pt}
  \textit{- Relevance}: Why did you choose the information you selected? How did it help answer the question?\\
 \textit{- Missed Insights}: Were there other relevant pieces of information you didn't consider? If so, why?\\
 \textit{ - Redundancies}: Did you include irrelevant or redundant information? If yes, identify and revise.
\end{adjustwidth}
\begin{adjustwidth}{6pt}{0pt}
\textbf{- Align the problem understanding and graph function information}
\end{adjustwidth}
\begin{adjustwidth}{11pt}{0pt}
  \textit{- Understanding and Selection}: How did your understanding of the problem influence your graph structure choice?\\
  \textit{- Inconsistencies}: Were there any mismatches between your understanding and your graph structure selection? If so, what caused them?\\
  \textit{- Adjustments}: How can you better align your understanding with the graph structure selection?
\end{adjustwidth}
\textbf{- Improved Strategy}
\begin{adjustwidth}{6pt}{0pt}
 Based on your reflection:\\
  \textbf{- Updated Understanding of the Problem}: Revise and describe your updated understanding.\\
  \textbf{- Revised Graph Selection}: List and explain which graph information you would now choose and why it is more suitable, where graph functions can be combined into compound functions to streamline operations. Ensure no more than two functions are combined at each step.\\
  \textbf{- Avoiding Past Issues}: Describe how this strategy addresses the challenges and improves your reasoning.
\end{adjustwidth}
\end{tcolorbox}

\begin{tcolorbox}[colback=white!10!white, colframe=blue!30!black, title=, boxrule=1pt]
\justifying
\noindent
Here are some examples:\\
\{\textit{examples}\}\\
(END OF EXAMPLES)\\
\textbf{Previous trial Details}:\\
Question: \{\textit{question}\} Please answer by providing node main feature (e.g., names) rather than node IDs.\\
\{\textit{scratchpad}\}\\
Reflection:
\end{tcolorbox}

\subsection{Model Evaluation Prompts}
\label{sec:appendixD3}
\begin{tcolorbox}[colback=white!10!white, colframe=blue!30!black, title=Evaluation Prompts, boxrule=1pt]
\justifying
\noindent
You are an intelligent reasoning accuracy evaluation agent. Evaluate the final answer based on all the plan, thought, action, and observation processes and determine if it meets the problem requirements. Ensure the following: The final answer directly corresponds to the data retrieved from the graph. It satisfies the question's requirement without including irrelevant or incorrect elements. The reasoning behind the answer is logical and supported by the observations. In a few sentences, please provide a brief explanation summarizing why the answer meets or does not meet the criteria. Then, please conclude with a clear judgment based on the  explanation, respond [yes] if the answer is correct , or [no] if the answer is not correct.\\
Here are some examples:\\
\{\textit{examples}\}\\
(END OF EXAMPLES)\\
Solve a question answering task with interleaving Thought, Interaction with Graph, Feedback from Graph steps. In Plan step, you can think about what the question is asking and plan how to do to get the answer. In Thought step, you can think about what further information is needed, and In Interaction step, you can get feedback from graphs with four functions: \\
(1) Retrieve[keyword], which retrieves the related node from the graph according to the corresponding query.\\
\mbox{}
\end{tcolorbox}

\begin{tcolorbox}[colback=white!10!white, colframe=blue!30!black, title=, boxrule=1pt]
\justifying
\noindent
(2) Feature[Node, feature], which returns the detailed attribute information of Node regarding the given "feature" key.\\
(3) Degree[Node, neighbor\_type], which calculates the number of "neighbor\_type" neighbors of the node Node in the graph.\\
(4) Neighbour[Node, neighbor\_type], which lists the "neighbor\_type" neighbours of the node Node in the graph and returns them.\\
Besides, you can use compound function, such as Feature[Retrieve [keyword], feature], 
which returns the detailed attribute information of Retrieve[keyword] regarding the given "feature" key.
When last Observation has been given or there is no Plan, you should provide next only one Plan based on the question. When last Plan has been given, you should provide next only one Thought. When last Thought has been given, you should provide next only one Action. When you think it's time to finish, use Finish to end the process. Don't make Observation.\\
Definition of the graph: \\\{\textit{graph}\_\textit{definition}\}\\
Question: \{\textit{question}\} Please answer by providing node main feature (e.g., names) rather than node IDs.\\\{\textit{scratchpad}\}\\
Proceed with explanation and judgment below:
\end{tcolorbox}

\end{document}